\documentclass[journal,onecolumn]{IEEEtran}
\ifCLASSINFOpdf
\else
\fi
\usepackage{epsfig}
\usepackage{graphicx}
\usepackage{amsmath}
\usepackage{amssymb}
\usepackage{commath}
\usepackage{bbding}
\usepackage{cite}
\usepackage{cleveref}
\usepackage{booktabs,multirow}

\usepackage{epstopdf}

\usepackage{amsmath,graphicx}
\usepackage{amsfonts}
\usepackage{subfigure}

\usepackage{bbm,booktabs,tabulary}

\usepackage{commath}
\usepackage[mathscr]{euscript}
\let\euscr\mathscr \let\mathscr\relax
\usepackage[scr]{rsfso}
\usepackage{booktabs}
\usepackage{mathtools}
\DeclarePairedDelimiter\ceil{\lceil}{\rceil}
\DeclarePairedDelimiter\floor{\lfloor}{\rfloor}

\begin{document}
%
\title{Motion-Compensated Temporal Filtering for Critically-Sampled Wavelet-Encoded Images}
%
%
%

\author{Vildan Atalay Aydin and Hassan Foroosh
\thanks{Vildan Atalay Aydin and Hassan Foroosh are with the Department of Computer Science, University of Central Florida, Orlando,
FL, 32816 USA (e-mail: vatalay@knights.ucf.edu, foroosh@cs.ucf.edu).}
}

\maketitle

\begin{abstract}
We propose a novel motion estimation/compensation (ME/MC) method for wavelet-based (in-band) motion compensated temporal filtering (MCTF), with application to low-bitrate video coding. Unlike the conventional in-band MCTF algorithms, which require redundancy to overcome the shift-variance problem of critically sampled (complete) discrete wavelet transforms (DWT), we perform ME/MC steps directly on DWT coefficients by avoiding the need of shift-invariance. We omit upsampling, the inverse-DWT (IDWT), and the calculation of redundant DWT coefficients, while achieving arbitrary subpixel accuracy without interpolation, and high video quality even at very low-bitrates, by deriving the exact relationships between DWT subbands of input image sequences. Experimental results demonstrate the accuracy of the proposed method, confirming that our model for ME/MC effectively improves video coding quality.
\end{abstract}



\section{Introduction}
\label{sec:intro}

Motion Estimation/Compensation and video coding have wide range of applications in various areas of image/video processing, including restoration \cite{Foroosh_2000,Foroosh_Chellappa_1999,Foroosh_etal_1996,Cao_etal_2015,berthod1994reconstruction,shekarforoush19953d,lorette1997super,shekarforoush1998multi,shekarforoush1996super,shekarforoush1995sub,shekarforoush1999conditioning,shekarforoush1998adaptive,berthod1994refining,shekarforoush1998denoising,bhutta2006blind,jain2008super,shekarforoush2000noise,shekarforoush1999super,shekarforoush1998blind,amiot2015fluorosocopic}, 
content/context analysis \cite{Tariq_etal_2017_2,Tariq_etal_2017,tariq2013exploiting,tariq2014scene,Cakmakci_etal_2008,Cakmakci_etal_2008_2,Zhang_etal_2015,Lotfian_Foroosh_2017,Morley_Foroosh2017,Ali-Foroosh2016,Ali-Foroosh2015,Einsele_Foroosh_2015,ali2016character,Cakmakci_etal2008,damkjer2014mesh}, surveillance  \cite{Junejo_etal_2007,Junejo_Foroosh_2008,Sun_etal_2012,junejo2007trajectory,sun2011motion,Ashraf_etal2012,sun2014feature,Junejo_Foroosh2007-1,Junejo_Foroosh2007-2,Junejo_Foroosh2007-3,Junejo_Foroosh2006-1,Junejo_Foroosh2006-2,ashraf2012motion,ashraf2015motion,sun2014should}, action recognition \cite{Shen_Foroosh_2009,Ashraf_etal_2014,Ashraf_etal_2013,Sun_etal_2015,shen2008view,sun2011action,ashraf2014view,shen2008action,shen2008view-2,ashraf2013view,ashraf2010view,boyraz122014action,Shen_Foroosh_FR2008,Shen_Foroosh_pose2008,ashraf2012human}, self-localization \cite{Junejo_etal_2010,Junejo_Foroosh_2010,Junejo_Foroosh_solar2008,Junejo_Foroosh_GPS2008,junejo2006calibrating,junejo2008gps}, tracking \cite{Shu_etal_2016,Milikan_etal_2017,Millikan_etal2015,shekarforoush2000multi,millikan2015initialized}, scene modeling \cite{Junejo_etal_2013,bhutta2011selective,junejo1dynamic,ashraf2007near}, and video post-production \cite{Cao_etal_2005,Cao_etal_2009,shen2006video,balci2006real,xiao20063d,moore2008learning,alnasser2006image,Alnasser_Foroosh_rend2006,fu2004expression,balci2006image,xiao2006new,cao2006synthesizing}. 
 to name a few.\\

Reliable motion estimation/compensation can substantially reduce the residual energy in the coding of video data. Motion estimation methods are either global  \cite{Foroosh_etal_2002,Foroosh_2005,Balci_Foroosh_2006,Balci_Foroosh_2006_2,Alnasser_Foroosh_2008,Atalay_Foroosh_2017,Atalay_Foroosh_2017-2,shekarforoush1996subpixel,foroosh2004sub,shekarforoush1995subpixel,balci2005inferring,balci2005estimating,foroosh2003motion,Balci_Foroosh_phase2005,Foroosh_Balci_2004,balci2006subpixel,balci2006alignment}, or local \cite{foroosh2001closed,shekarforoush2000multifractal,foroosh2004adaptive,foroosh2003adaptive} in their nature in terms of treating the transformation relating two images. There is also a separate but related body of work on camera motion quantification, which requires online or offline calibration of camera \cite{Cao_Foroosh_2007,Cao_Foroosh_2006,Cao_etal_2006,Junejo_etal_2011,cao2004camera,cao2004simple,caometrology,junejo2006dissecting,junejo2007robust,cao2006self,foroosh2005self,junejo2006robust,Junejo_Foroosh_calib2008,Junejo_Foroosh_PTZ2008,Junejo_Foroosh_SolCalib2008,Ashraf_Foroosh_2008,Junejo_Foroosh_Givens2008,Lu_Foroosh2006,Balci_Foroosh_metro2005,Cao_Foroosh_calib2004,Cao_Foroosh_calib2004,cao2006camera}.
While these methods and their variations have been proposed in the past for motion compensation in different applications, space-time subband/wavelet coding \cite{ohm2005advances} is by far the method of choice for coding and compressing images and videos due to its superior performance. Its effectiveness, however, can be significantly improved with motion compensation, which is the topic of the proposed method in this paper. 

\section{Related Work}
\label{sec:related}
Still image coding \cite{andreopoulos2005complete} and video coding \cite{XuLW15} are important topics of research in coding and compression of multimedia data. On the other hand, scalable video coding \cite{van2002bottom,park2000motion} is an emerging trend in numerous multimedia applications with heterogeneous networks, due to their ability to adapt different resolution and quality requirements. Recently, a large body of research has focused on wavelet-based methods \cite{secker2001motion,andreopoulos2005complete,liu2007fast,chen2014adaptive}, where motion compensated temporal filtering (MCTF) is shown to play an essential role in both scalable video coding and still image coding. MCTF is performed either directly on input images, or on their transforms. Thus, MCTF methods can be categorized into two groups depending on the order of temporal and spatial transforms. MCTF techniques which perform temporal decomposition before a spatial transform include, Secker and Taubman \cite{secker2001motion}, and Pesquest-Popescu and Bottreau \cite{pesquet2001three} who used lifting formulation of three dimensional temporal wavelet decomposition for motion compensated video compression. Kim \textit{et al.} \cite{kim2000low} proposed a 3-D extension of set partitioning in hierarchical trees (3D-SPIHT) by a low bit-rate embedded video coding scheme. More recently, Xiong \textit{et al.} \cite{xiong2008scale} extended spatiotemporal subband transform to in-scale motion compensation to exploit the temporal and cross-resolution correlations simultaneously, by predicting low-pass subbands from next lower resolution and high-pass subbands from neighboring frames in the same resolution layer. Furthermore, Chen and Liu \cite{chen2014adaptive} used an adaptive Lagrange multiplier selection model in rate-distortion optimization (RDO) for motion estimation. In order to achieve more accurate motion data, Esche \textit{et al.} \cite{esche2013adpative} proposed an interpolation method for motion information per pixel using block based motion data, and R{\"u}fenacht \textit{et al.} \cite{rufenacht2014hierarchical} anchor motion fields at reference frames instead of target frames to resolve folding ambiguities in the vicinity of motion discontinuities. 

Although the methods cited above have good performance, they suffer from drifting and operational mismatch problems \cite{xiong2008scale}. Therefore, performing spatial transform before temporal decomposition was introduced to overcome these drawbacks. However, since complete DWT is shift variant, in order to achieve in-band ME/MC (i.e. directly in the wavelet domain), several methods were proposed to tackle this problem by redundancy. Van der Auwera \textit{et al.} \cite{van2002bottom} used a bottom-up prediction algorithm for a bottom-up overcomplete discrete wavelet transform (ODWT). Park and Kim \cite{park2000motion} proposed a low-band-shift method by constructing the wavelet tree by shifting low-band subband in each level for horizontal, vertical, and diagonal directions for one pixel and performing downsampling. Andreopoulos \textit{et al.} \cite{andreopoulos2005complete} defined a complete to overcomplete discrete wavelet transform (CODWT), which avoids inverse DWT generally used to obtain ODWT. More recently, Liu and Ngan \cite{liu2007fast} use partial distortion search and anisotropic double cross search algorithms with the MCTF method in \cite{andreopoulos2005complete} for a fast motion estimation. Amiot \textit{et al.} \cite{amiot2015fluorosocopic} perform MCTF for denoising, using dual-tree complex wavelet (DT-CW) coefficients. 

All MCTF methods summarized above perform motion estimation/motion compensation either in the temporal domain before DWT, or in the wavelet domain with the help of redundancy (e.g. ODWT, DT-CW,  etc.), due to the fact that complete DWT is shift-variant and motion estimation directly on DWT subbands is a challenging task. However, redundancy in these methods leads to high computational complexity \cite{liu2007fast}. Inspired by the fact that shift variance keeps the perfect reconstruction and nonredundancy properties of wavelets and breaks the coupling between spatial subbands, and that wavelet codecs always operate on complete DWT subbands \cite{andreopoulos2005complete}, we propose a novel in-band ME/MC method, which avoids the need of shift invariance, and operates directly on the original DWT coefficients of the input sequences. Since Haar wavelets are widely utilized in MCTF methods due to the coding efficiency based on their short kernel filters \cite{andreopoulos2005complete}, our method is built on Haar subbands. For accurate ME/MC, we define the exact relationships between the DWT subbands of input video sequences, which allows us to avoid upsampling, inverse DWT, redundancy, and interpolation for subpixel accuracy. 

The rest of the paper is organized as follows. We introduce the problem and our proposed solution in Section \ref{sec:formulation}. We define the derived exact inter-subband relationships in Section \ref{sec:method}, demonstrate the experimental results in Section \ref{sec:results}, and finally conclude our paper in Section \ref{sec:conclusion}.

\section{Motion Compensated Temporal Filtering}
\label{sec:formulation}
In this section, we explain our proposed method for in-band motion compensated temporal filtering, operating directly on DWT subbands. 


\begin{figure}[t]
  \centerline{\includegraphics[width=12cm]{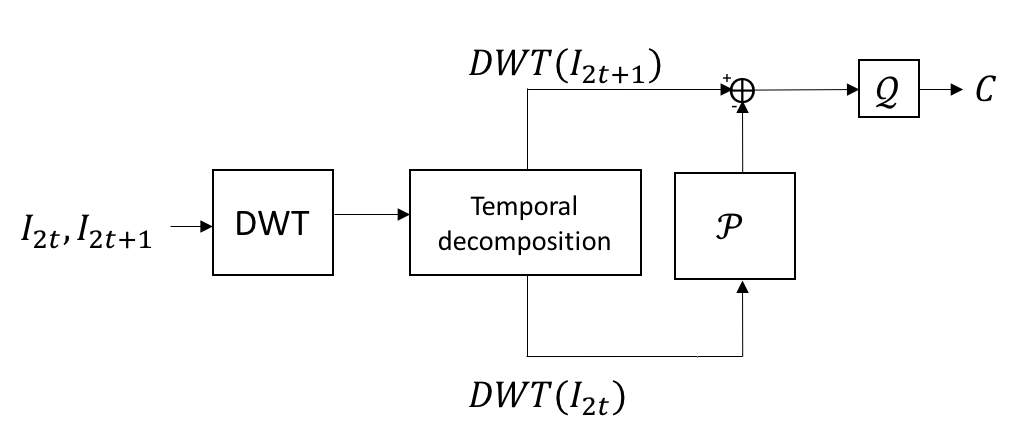}}
\centering
  \caption{A block diagram of the proposed in-band Motion Compensated Temporal Filtering model.}\medskip \label{fig:MCTF}
\end{figure}

The wavelet transform provides localization both in time and frequency; therefore, it is straightforward to use wavelets in MCTF. In order to perform ME/MC in MCTF, wavelet subbands of the transformed signal need to be predicted. However, due to decimation and expansion operations of DWT, direct band-to-band estimation is generally not practical \cite{park2000motion}. The proposed method overcomes this challenge by revealing the relationships between subbands of reference and target frames.


The proposed in-band MCTF method is demonstrated in Fig. \ref{fig:MCTF}. Given a video sequence, first, DWT is performed on each frame for spatial decomposition, then a temporal decomposition is performed by splitting video frames into groups. ME/MC ($\euscr{P}$ in Fig. \ref{fig:MCTF})  is performed by block matching, using reference frames ($DWT(I_{2t})$) to predict the target frames ($DWT(I_{2t+1})$). Employing the found motion vectors (MV), reference frames are mapped onto the target frames to generate error frames, $C$ in Fig. \ref{fig:MCTF}, which are then quantized ($\euscr{Q}$), encoded/decoded by a wavelet codec, together with the MVs. 

We employ Haar wavelet decomposition in spatial transform due to the benefits mentioned earlier. Since the method in Section \ref{sec:method} is accurate for any arbitrary subpixel translation defined as a multiple of $2^k$, where $k$ is the decomposition level, our method does not need interpolation for subpixel accuracy. A block matching method with unidirectional full search is used for ME/MC steps which is a common method used for MCTF. Our cost function is based on mean square error minimization using all subbands, as follows:

\begin{equation}
(dx,dy) = \operatorname{arg\,min}_{x,y} \{( A - \hat{A} )^{2} + ( a - \hat{a} )^{2} + ( b - \hat{b} )^{2} + ( c - \hat{c} )^{2}\},
\end{equation}

\noindent where $A, a, b, c$ denote the original target frame wavelet subbands, and $\hat{A}, \hat{a}, \hat{b}, \hat{c}$ are the estimated subbands for the same target image, using the method described in Section \ref{sec:method} and a reference frame.

\section{Inter-subband Relationship}
\label{sec:method}
In-band (wavelet domain) shift method along with the related notation are provided in this section.

\subsection{Notation} \label{term}

Here, we provide the notation used throughout the paper beforehand, in Table \ref{termtable}, for a better understanding of the proposed method and to prevent any confusions. 
\begin{center}
	\begin{table}[h] 
		\centering
		\caption{Notation}
		\begin{tabular}{l p{0.7\linewidth}}
			$I_t$ & Input video frame at time $t$\\
			${\bf A}, {\bf a}, {\bf b}, {\bf c}$ & Haar wavelet transform approximation, horizontal, vertical, and diagonal subbands of input image, respectively \\
			${\bf F}, {\bf K}, {\bf L}$ & Coefficient matrices to be multiplied by approximation, horizontal, vertical, and diagonal DWT subbands \\
			$h$ & Number of hypothetically added levels in case of non-integer shifts\\
			$s$ & Integer shift amount after the hypothetically added levels ($h$)\\
		\end{tabular} \label{termtable}
	\end{table} 
\end{center}

Bold letters in the following sections demonstrate matrices and vectors. The subscripts $x$ and $y$ indicate the horizontal and vertical translation directions, respectively. Finally, the subscript $k$ indicates the $k$th video frame, where $k = 1, 2, ..$.

\subsection{In-band Shifts}
Our goal for the MCTF method described in Section \ref{sec:formulation} is to achieve ME/MC in the wavelet domain using DWT subbands, given a video frame sequence. For this purpose, wavelet subbands of the tranformed signal should be predicted using only DWT subbands of the reference frame. Therefore, we derive the relationship between the subbands of transformed and reference images, which can be described by in-band shift (in the wavelet domain) of the reference image subbands. Below, we derive the mathematical expressions which demonstrate these relationships.

Let $A$, $a$, $b$, and $c$ be the first level approximation, horizontal, vertical, and diagonal detail coefficients (subbands), respectively, of a $2D$ reference frame at time $t$, $I_t$, of size $2m\times2n$, where $m$ and $n$ are positive integers. Since decimation operator in wavelet transform reduces the size of input frame by half in each direction for each subband, we require the frame sizes to be divisible by 2. Now, the $1st$ level subbands of translated frame in any direction (i.e. horizontal, vertical, diagonal) can be expressed in matrix form using the $1st$ level Haar transform subbands of reference frame as in the following equations:

\begin{eqnarray}\label{firsteq} 
\textbf{A}_s &=& \textbf{F}_y \textbf{A} \textbf{F}_x + \textbf{F}_y \textbf{a} \textbf{K}_1 + \textbf{L}_1 \textbf{b} \textbf{F}_x + \textbf{L}_1 \textbf{c} \textbf{K}_1 \nonumber\\ 
\textbf{a}_s &=& - \textbf{F}_y \textbf{A} \textbf{K}_1 + \textbf{F}_y \textbf{a} \textbf{K}_2 - \textbf{L}_1 \textbf{b} \textbf{K}_1 + \textbf{L}_1 \textbf{c} \textbf{K}_2 \nonumber\\
\textbf{b}_s &=& - \textbf{L}_1 \textbf{A} \textbf{F}_x - \textbf{L}_1 \textbf{a} \textbf{K}_1 + \textbf{L}_2 \textbf{b} \textbf{F}_x + \textbf{L}_2 \textbf{c} \textbf{K}_1 \nonumber\\
\textbf{c}_s &=& \textbf{L}_1 \textbf{A} \textbf{K}_1 - \textbf{L}_1 \textbf{a} \textbf{K}_2 - \textbf{L}_2 \textbf{b} \textbf{K}_1 + \textbf{L}_2 \textbf{c} \textbf{K}_2 \nonumber\\
\end{eqnarray} 

As already mentioned in Section \ref{term}, $\textbf{F}$, $\textbf{K}$, and $\textbf{L}$ stand for coefficient matrices to be multiplied by the lowpass and highpass subbands of the reference frame, where subscripts $x$ and $y$ indicate \textit{horizontal} and \textit{vertical} shifts. $\textbf{A}_s, \textbf{a}_s, \textbf{b}_s, \textbf{c}_s$ are translated frame subbands in any direction. The low/high-pass subbands of both reference and transformed frames are of size $m \times n$, $\textbf{F}_y$ and $\textbf{L}_{1,2}$ are $m \times m$, whereas $\textbf{F}_x$ and $\textbf{K}_{1,2}$ are $n \times n$.

By examining the translational shifts between subbands of two input frames in the Haar domain, we realize that horizontal translation reduces $\textbf{L}$ to zero and $\textbf{F}_y$ to the identity matrix. This could be understood by examining the coefficient matrices defined later in this section (namely, Eq. (\ref{coefmat})), by setting the related vertical components to zero (specifically, $s_y$ and $h_y$). Likewise, vertical translation depends solely on approximation and vertical detail coefficients, in which case $\textbf{K}$ is reduced to zero and $\textbf{F}_x$ is equal to the identity matrix. 

Here, we first define the matrices for subpixel shift amounts. The algorithm to reach any shift amount using the subpixel relationship will be described later in this section.

\begin{figure}[t]
	\centering
	\centerline{\includegraphics[width=6.5cm]{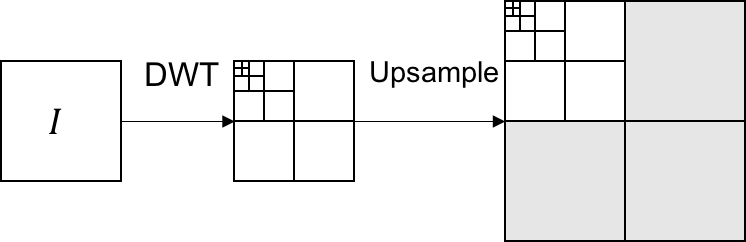}}
	\caption{Upsampling illustration.}\medskip \label{fig:upsample}
\end{figure}

For subpixel translation, contrary to the customary model of approximating a subpixel shift by upsampling an image followed by an integer shift, our method models subpixel shift directly based on the original coefficients of the reference frame, without upsampling and the need for interpolation. To this end, we resort to the following observations: 

\textbf{(1)} Upsampling an image $I$, is equivalent to adding levels to the bottom of its wavelet transform, and setting the detail coefficients to zero, while the approximation coefficients remain the same, as demonstrated in Fig. \ref{fig:upsample} for upsampling by $2^1$ as an example, where gray subbands show added zeros. 

\textbf{(2)} Shifting the upsampled image by an amount of $s$ is equivalent to shifting the original image by an amount of $s/2^h$, where $h$ is the number of added levels (e.g. $h=1$ in Fig. \ref{fig:upsample}). 

These observations allow us to do an in-band shift of the reference subbands for a subpixel amount, without upsampling or interpolation, which saves both memory and reduces the computational cost. Transformed signals therefore can be found by using the original level subbands of the reference image with the help of a hypothetically added level ($h$) and an integer shift value ($s$) at the added level.

Now, the aforementioned coefficient matrices, $\textbf{F}_x$, $\textbf{K}_{1}$, and $\textbf{K}_{2}$ can be defined, in lower bidiagonal Toeplitz matrix form as follows.

\scriptsize
\begin{eqnarray} 
\resizebox{\linewidth}{!}{%
	$\textbf{F}_x = \dfrac{1}{2^{h_x+1}} 
	\begin{bmatrix}
	2^{h_x+1} - \abs{s_x} & &   \\
	\abs{s_x} & 2^{h_x+1} - \abs{s_x}   \\
	& \abs{s_x} \\
	& & \ddots & \ddots \\
	& \\
	& & & \abs{s_x} & 2^{h_x+1} - \abs{s_x} \\ \nonumber
	\end{bmatrix} \nonumber$
}
\end{eqnarray} 


\begin{eqnarray}
\textbf{K}_1 = \dfrac{1}{2^{h_x+1}} 
\begin{bmatrix}
-s_x &  \\
s_x & -s_x   \\
& s_x & \\
& & \ddots & \ddots\\
&\\
& & & s_x & -s_x \\ 
\end{bmatrix} \nonumber
\end{eqnarray}


\begin{eqnarray}
\resizebox{\linewidth}{!}{%
$\textbf{K}_2 = \dfrac{1}{2^{h_x+1}} 
\begin{bmatrix}
2^{h_x+1} - 3\abs{s_x} &  \\
- \abs{s_x} & 2^{h_x+1} - 3\abs{s_x}   \\
& -\abs{s_x}   & \\
& & \ddots & \ddots\\
&\\
& & & -\abs{s_x} & 2^{h_x+1} - 3\abs{s_x}  \\ 
\end{bmatrix}$ \label{coefmat}
}
\end{eqnarray}


\normalsize
\noindent where $s_{x}$ and $h_{x}$ demonstrate the integer shift amounts at the hypothetically added level and the number of added levels for $x$ direction, respectively. 

$\textbf{F}_y$, $\textbf{L}_1$, and $\textbf{L}_2$ matrices are defined in a similar manner by upper bidiagonal Toeplitz matrices, using $y$ direction values for $s$ and $h$.

As mentioned earlier, $\textbf{F}_x$ and $\textbf{K}_{1,2}$ are $n \times n$, while $\textbf{F}_y$ and $\textbf{L}_{1,2}$ are $m \times m$. Sizes of these matrices also indicate that in-band shift of subbands is performed using only the original level Haar coefficients (which are of size $m \times n$) without upsampling. When the shift amount is negative, diagonals of the coefficient matrices interchange. The matrices are adapted for boundary condition by adding one more column/row at the end, for the MCTF method proposed in Section \ref{sec:formulation}, where subband sizes are also adjusted to be $(m+1)\times(n+1)$.

The relationship defined above for subpixel shifts, can be used to produce any shift amount based on the fact that wavelet subbands are periodically shift-invariant. Table \ref{shifts} demonstrates the calculation of any shift using subpixels, where $\%$ stands for modulo, $\floor{s}$ and $\ceil{s}$ are the greatest integer lower than, and smallest integer higher than the shift amount $s$. Using circular shift of subbands for the given amounts in each shift amount case, and setting the new shift amount to the new shift values in Table \ref{shifts}, we can calculate any fractional or integer amount of shifts using subpixels.

\begin{center}
	\begin{table}[h] 
		\centering
		\caption{Arbitrary shifts defined by circular shift and subpixel amount}
		\begin{tabular}{lll}
			\toprule
			Shift amount & Circular shift & New shift amount\\
			\midrule
			$s\%2 = 0$ & $s/2$ 	& $0$\\	
			$s\%2 = 1$ & $\floor{s/2}$ & $1$ \\
			$\ceil{s}\%2 = 0$ & $\ceil{s}/2$ & $s-\ceil{s}$ \\
			$\floor{s}\%2 = 0$ & $\floor{s}/2$ & $s-\floor{s}$ \\
			\bottomrule	
		\end{tabular} \label{shifts}
	\end{table} 
\end{center}

If the shift amount (or the new shift amount in Table \ref{shifts}) is not divisible by $2$, in order to reach an integer value at the $(N+h)$th level, the shift value at the original level is rounded to the closest decimal point which is divisible by $2^h$.

\section{Experimental Results}
\label{sec:results}
In this section, we demonstrate the results obtained with our method compared to the methods which perform in-band MCTF for video coding. We report our results on CIF video sequence examples with resolutions $352\times240$ and $352\times288$. We set our block size to $8\times8$ or $16\times16$ depending on the resolution of the sequences (in order to have integer number of blocks in subbands) and the required accuracy. Even though our MCTF method is based on 1-level DWT, we perform $2$ more spatial decomposition levels after ME/MC steps before encoding, since compared methods use $3$ spatial decomposition levels in total. Motion vectors and error frames are encoded using context-adaptive variable-length coding (CAVLC) and 
global thresholding with Huffman coding methods, respectively. 

\begin{figure}[h]
	\centering
		\includegraphics[width=8cm]{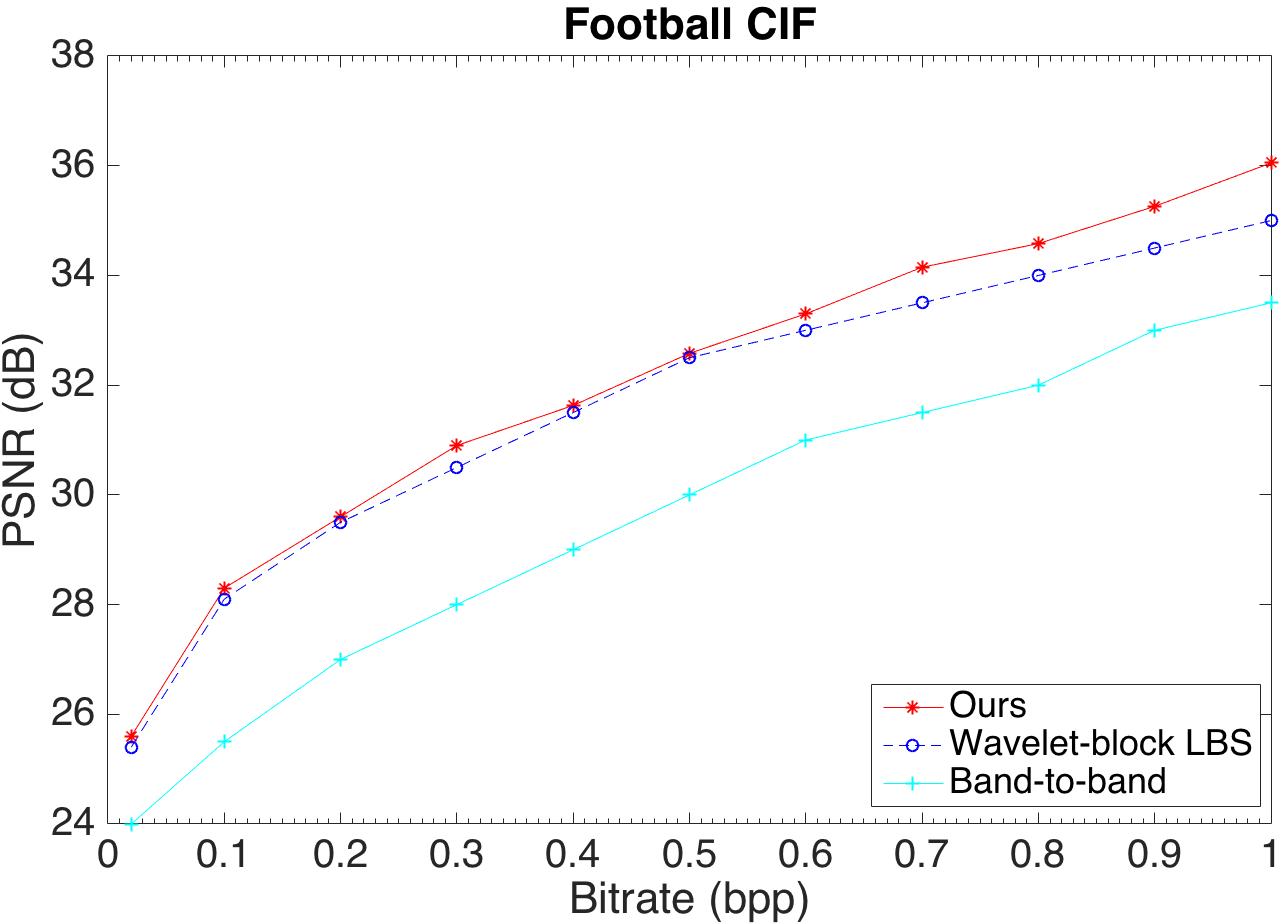}
			\caption{Rate-distortion comparison for the Football sequence.} \label{fig:football}
\end{figure}

Fig. \ref{fig:football} shows the comparison of our method with respect to two conventional in-band methods, which are direct wavelet subband matching (band-to-band) and wavelet-block low-band-shift (LBS) \cite{park2000motion} for CIF video sequence "Football". The graph demonstrates rate-distortion curves for a predicted frame of the Football sequence, where the shown bitrates are for error frame only (same as in the compared methods), and the accuracy for our method is set to $1/4$ pixel. As seen in this figure, our method improves PSNR compared to conventional in-band methods by $0.1-1$ dB in general.

\begin{figure}[h]
\begin{tabular}{cc}
		\includegraphics[width=7.7cm]{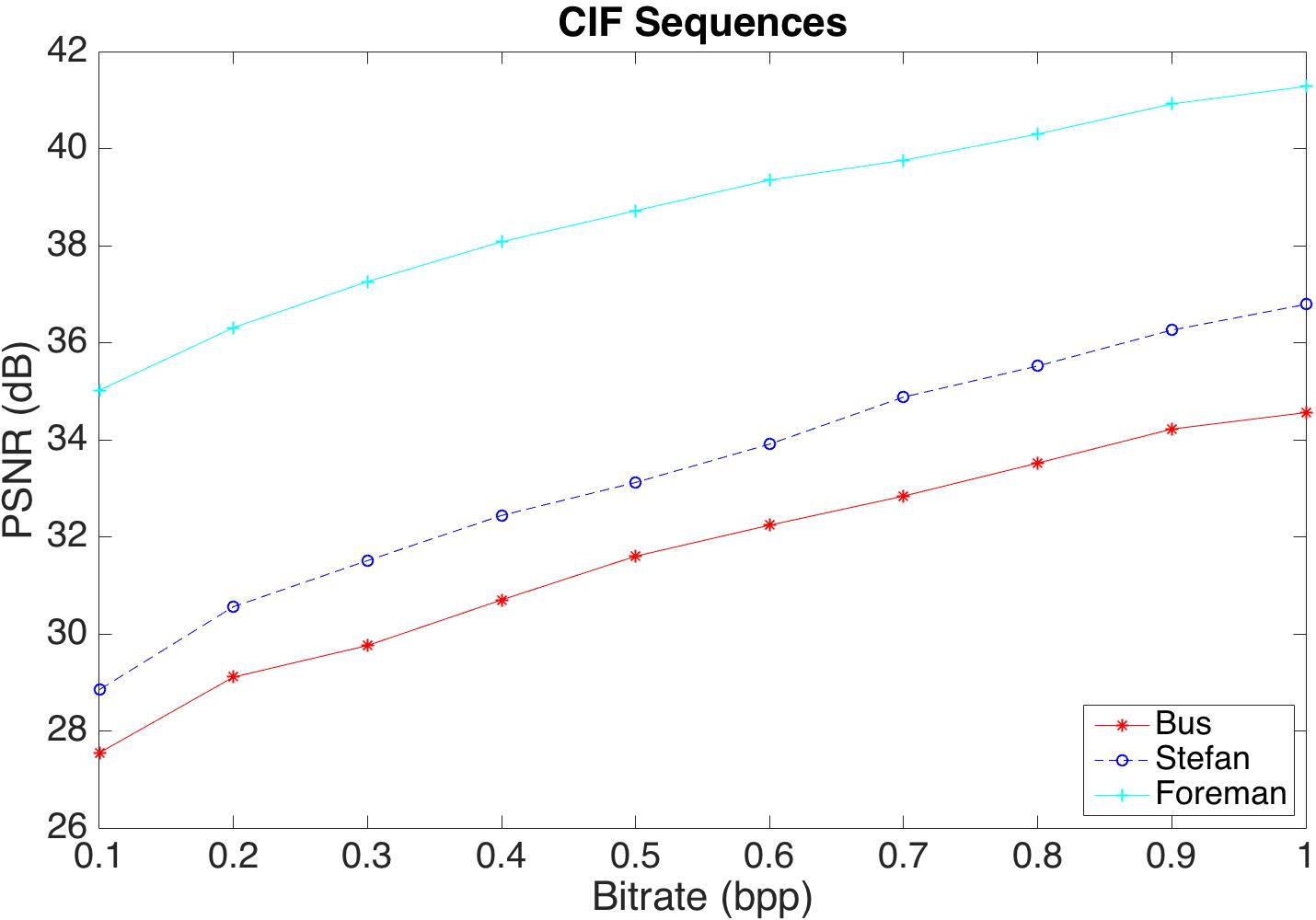} &
		\includegraphics[width=7.7cm]{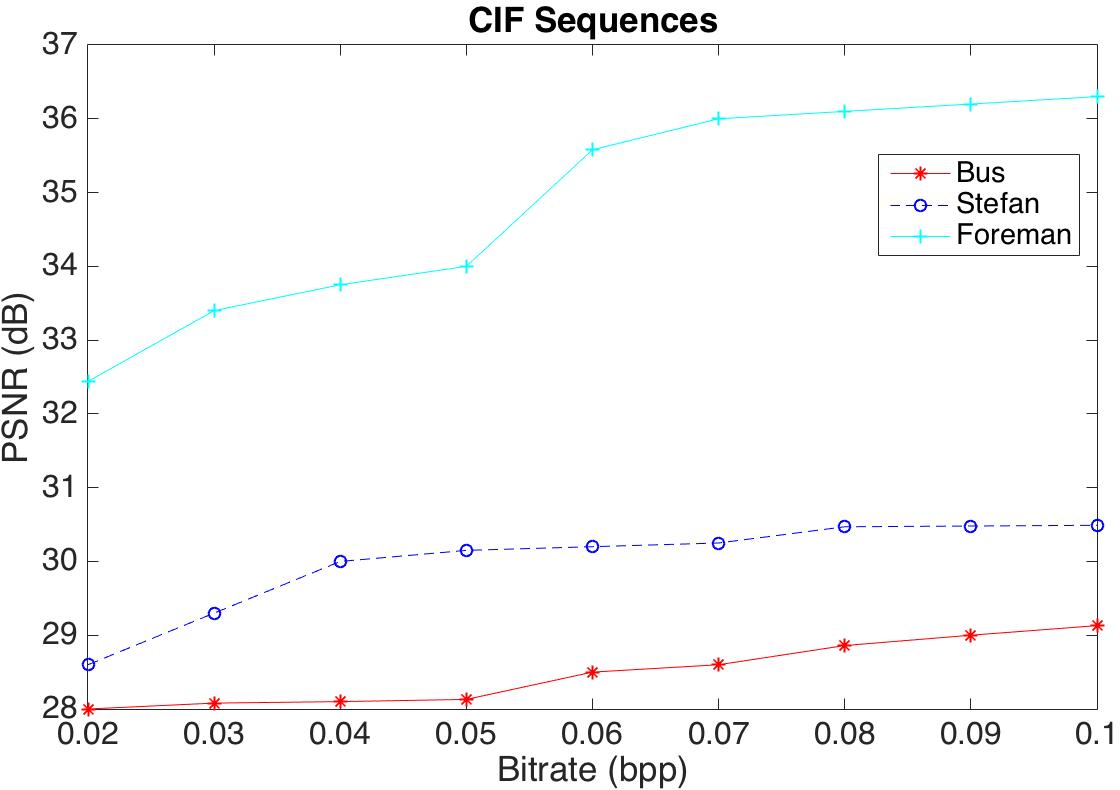}
\end{tabular}
	\caption{PSNR performance of proposed method.} \label{fig:ours}
\end{figure}

\begin{figure}[h]
	\centering
\begin{tabular}{cc}
		\includegraphics[width=8.3cm, height = 6cm]{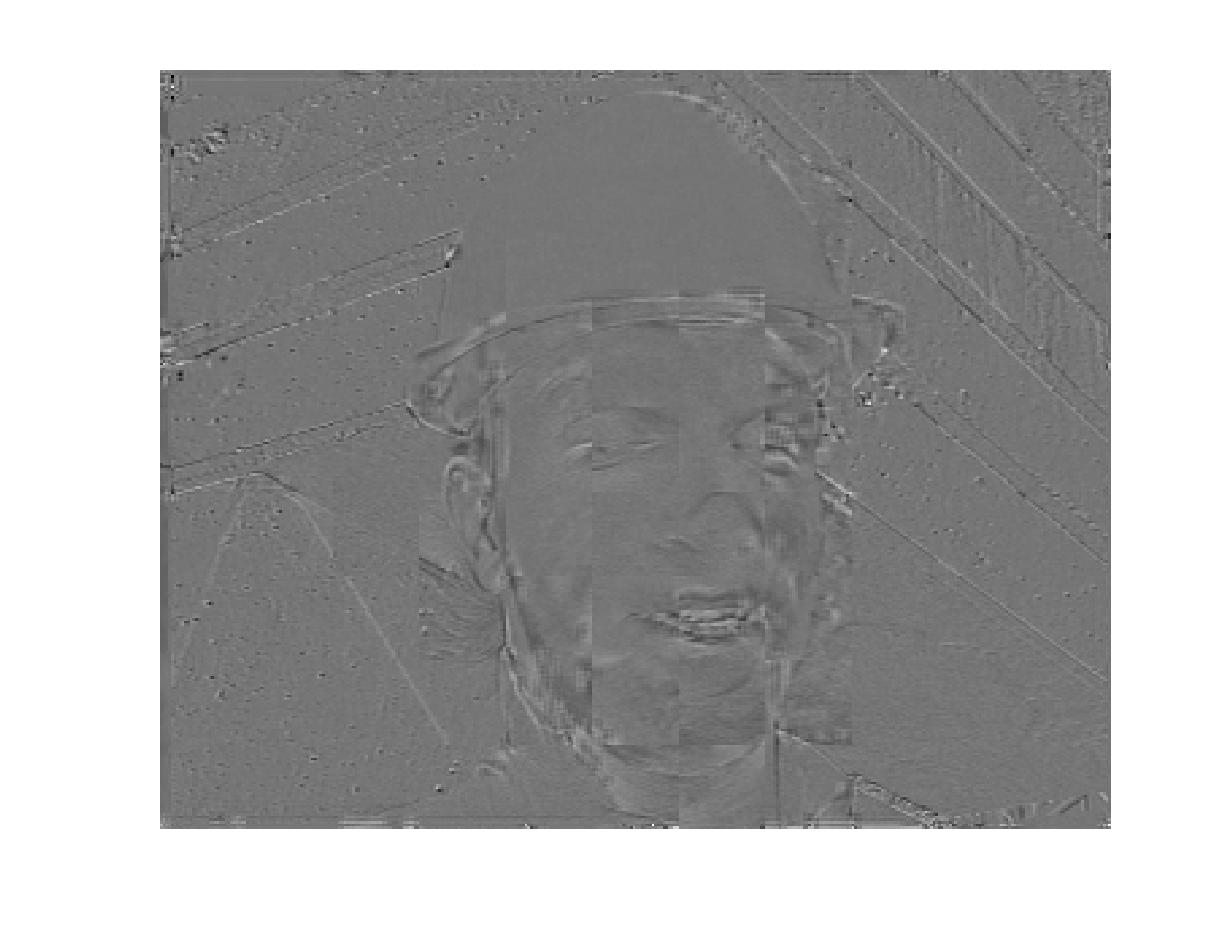}
		\includegraphics[width=8.3cm, height = 6cm]{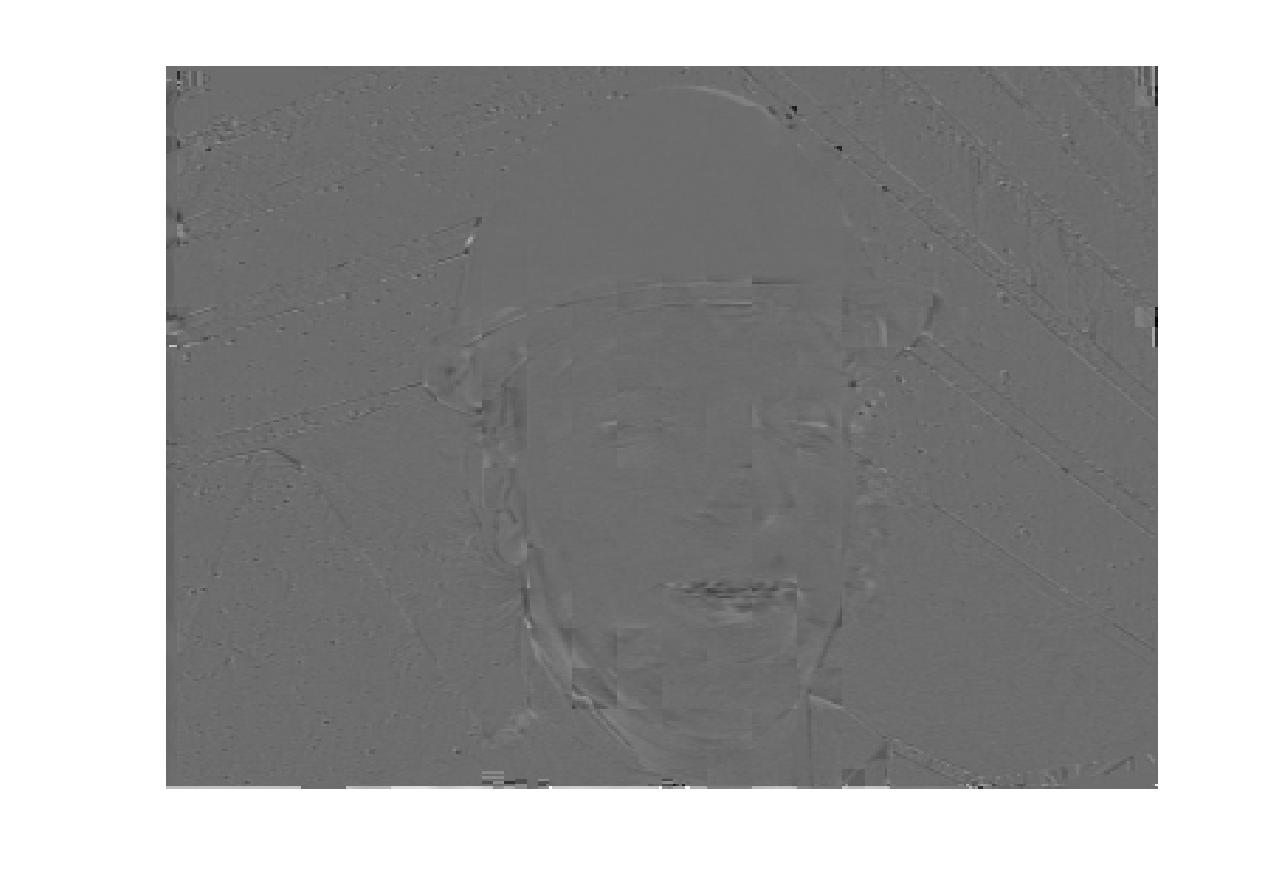}
\end{tabular}
	\caption{Residual images for predicted frames of Foreman for $0.1$ bpp on the left and $0.02$ bpp on the right.} \label{fig:residual}
\end{figure}

We demonstrate our results for several video sequences at different bitrates in Fig. \ref{fig:ours}, where bitrates include the luminance component only for the reference frame, the error frame, and MVs. The graph on the left shows the results with $1/2$ pixel accuracy using $16\times16$ blocks, and the one on the right uses $1/4$ pixel accuracy with $8\times8$ blocks. We also show the residual images for a predicted frame of the Foreman sequence in Fig. \ref{fig:residual}, for $0.1$ and $0.02$ bpp, respectively. The examples show how our method reduces the residual signal energy even at very low bitrates by providing more accurate reconstruction (prediction).

\section{Conclusion}
\label{sec:conclusion}
We propose a novel method for wavelet-based (in-band) ME/MC for MCTF in for video coding, where DWT is applied before temporal decomposition, and ME/MC steps are performed directly on DWT subbands. We avoid the need for shift-invariance property for non-redundant DWT (required by conventional methods for ME/MC), by deriving the exact relationships between DWT subbands of reference and transformed video frames.
Our method avoids upsampling, inverse-DWT (IDWT), and calculation of redundant DWT while achieving high accuracy even at very low-bitrates. Experimental results demonstrate the accuracy of presented method for ME/MC, confirming that our model effectively improves video coding quality by reducing the residual energy in the error frames. The proposed ME/MC scheme can also be adapted for several image/video processing applications such as denoising, or scalable video coding.

{
\bibliographystyle{plain}
\bibliography{strings,refs,foroosh}
}

\end{document}